\documentclass[letterpaper, 10pt, conference]{ieeeconf}
\IEEEoverridecommandlockouts
\usepackage{amsmath,amssymb,amsfonts}
\usepackage{graphicx}
\usepackage{booktabs}
\usepackage{url}
\usepackage{xcolor}
\usepackage{hyperref}
\usepackage{cuted}
\usepackage{capt-of}

\title{\LARGE \bf To Select or not to Select, that is the Question:\\Distilling Robot Skill Prediction into a Small Ensemble}
\author{Haechan Mark Bong$^{1, 2}$,
        Simon Roy$^{1, 2}$,
        Euhid Aman$^{3}$,
        Giovanni Beltrame$^{1, 2}$
        \thanks{$^{1}$Department of Computer Engineering and Software Engineering, Polytechnique Montréal, QC., Canada. {\tt\small haechan.bong@etud.polymtl.ca}}
        \thanks{$^{2}$MILA, QC., Canada.}
        \thanks{$^{3}$National Taiwan University of Science and Technology (NTUST), Taipei City, Taiwan.}}

\begin{document}
\bstctlcite{bstctl:etal}
\maketitle
\thispagestyle{empty}
\pagestyle{empty}
%\author{Anonymous Authors\\Anonymous Affiliation}

\begin{abstract}
As robot fleets become more heterogeneous, including humanoids, rovers,
quadrupeds, and drones, selecting the right robot for a task becomes a
core systems problem. We study \emph{robot skill prediction}: mapping a
natural-language task description to the physical capabilities required to
execute it, such as fly, wheels, legs, surface water, under water and
hands. Since labelled data that maps natural-language task descriptions to robot's physical capabilities does not exist, we construct a
synthetic task-to-skill dataset using LLM-assisted generation and targeted
label auditing. Trained on this data, a $\approx$133M-parameter ensemble of two
fine-tuned sentence encoders (mpnet + MiniLM) reaches 83.5\% task-to-skill matching
on a stratified 200 task dataset, outperforming Kimi K2 (1T MoE) at
72.0\%, GPT-OSS-120B at 71.5\%, and Llama-4-Scout-17B at 69.0\% under the same zero-shot prompt. These results suggest that, for fixed robot
skill taxonomies, small specialized models trained on synthetic data can
outperform much larger general-purpose LLMs for fleet-level task routing.
\end{abstract}

\section{Introduction}
\label{sec:intro}
To improve the integration of robots into our society, deploying them should
be as easy as issuing natural-language task instructions. As
heterogeneous robot fleets grow, a natural-language interface (e.g.,
``inspect the underside of the bridge for cracks'') must route each task
to a robot capable of executing it. A common decomposition
is \emph{skill prediction}: from the task description, predict which
physical skills (e.g., \texttt{fly}, \texttt{legs}, \texttt{wheels}, \texttt{under water}, \texttt{surface water} and
\texttt{hands}) are
required; matching those skills to available robots is then a trivial
lookup. This mapping is a difficult reasoning problem because a task may
require a single skill, several skills used together, or one of multiple
valid skill options. Accurate skill prediction decouples the language
problem from the fleet definition, producing a stable, robot-agnostic
multi-label classifier.

\begin{figure}[t]
\centering
\includegraphics[width=1\linewidth]{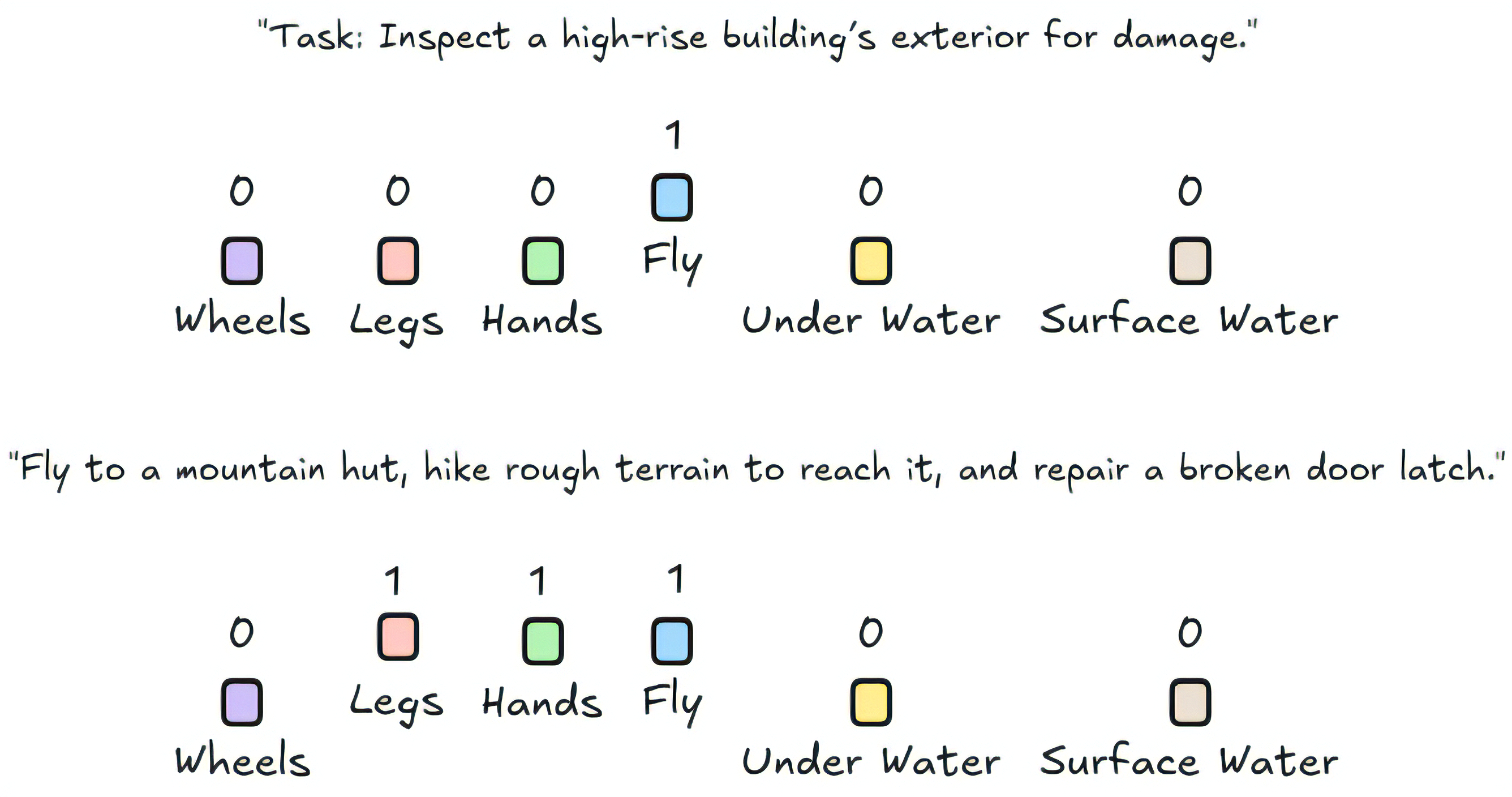}
\caption{Example of a single synthetic data with robot's phyisical capabilities (skills) as a boolean vector.}
\label{fig:data-example}
\end{figure}

The default approach would be to prompt a Large Language Model (LLM) using zero-shot learning: no training
data, no fine-tuning, no infrastructure. We show the opposite
trade-off is preferable: with a modest synthetic dataset and a small
fine-tuned ensemble, we outperform frontier LLMs (up to 1T parameters)
by atlest 11.5\% exact match (all and only required skill(s) are corrected predicted) while running locally at a fraction of
the compute. As a result, we highlight two main contributions:
\begin{enumerate}
    \item A \emph{synthetic task-description to robot skill matching dataset}
    with multi-label annotations spanning diverse application domains (figure~\ref{fig:data-example}).
    \item A lightweight \emph{small-ensemble model} trained on this dataset,
    showing that a specialized model can outperform much larger general-purpose
    LLMs on robot skill prediction.
\end{enumerate}

\section{Related Work}
\label{sec:related}
\textbf{Synthetic data for robot learning.} LLMs are increasingly used to generate training data for downstream robot and language tasks when manual data collection is expensive. Prior work assumes a robot is already selected and covers synthetic trajectories for manipulation and simulated demonstrations for imitation learning~\cite{gensim,mimicgen}, as well as language grounding for robot instruction following and task planning~\cite{blukis21a,sayplan}. Our focus at the top of the execution: learn to select the optimal robot based on natural lanaguage task description by predicting physical capabilities (skills) of robot types.

\textbf{Sentence encoders for classification.} Sentence
transformers~\cite{sentencetransformers} and their compressed
variants~\cite{mpnet, minilm} are commonly used for
lightweight semantic classification. Full fine-tuning on small
datasets risks overfitting, and we implemented a partial unfreezing of the top
transformer layers plus a classifier head to avoid this problem.

\textbf{LLMs for robot selection and task allocation.} To the best of our
knowledge, prior work does not directly study our setting of mapping a natural language task description to the robot's physical capabilties as skill set needed for robot
selection, which is at a morphological layer. Existing LLM-based robotics work is adjacent, but different: it focuses on task planning based on execution as skills, coalition formation, or task
allocation based on available sensors~\cite{smartllm, prieto2024taskalloc}, rather than fixed-taxonomy
multi-label skill prediction. This distinction matters because our goal is
not to generate a full plan or schedule, but to infer the minimal set of
skills required based on a task description to achieve optimal robot selection.

\begin{figure*}[!t]
\centering
\includegraphics[width=\textwidth]{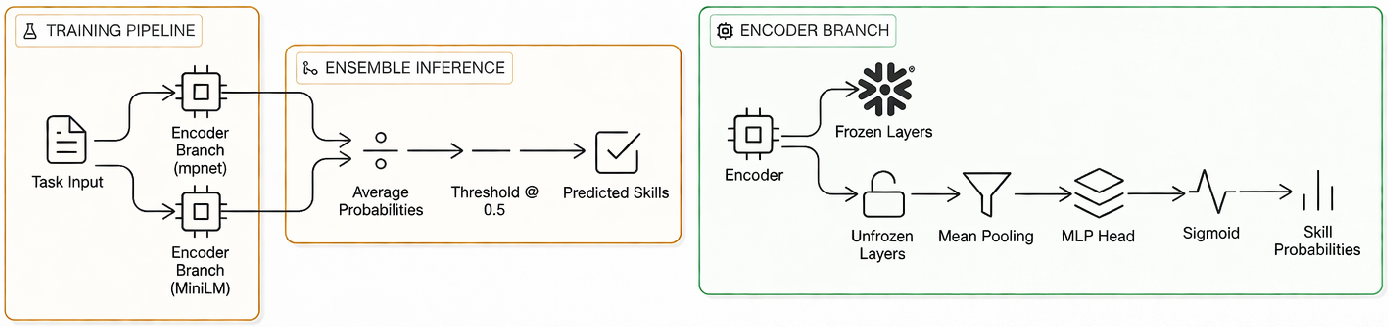}
\caption{Training and inference pipeline of the ensemble model.}
\label{fig:arch}
\end{figure*}

\section{Method}
\label{sec:method}

\subsection{Task Formulation}
\label{sec:task}

Each task description is assigned a binary vector over the skill set:
\begin{equation}
\small
\mathcal{S} = \left\{\begin{array}{c}
\texttt{fly},\, \texttt{legs},\, \texttt{wheels},
\texttt{hands},\\ \texttt{under water},\, \texttt{surface water}
\end{array}\right\}
\normalsize
\end{equation}

Skills are \emph{differentiating capabilities}, meaning capabilities that separate
one robot class from another, so universal capabilities (cameras,
GPS, onboard compute) are excluded by design. Prediction is multi-label:
a single task can require any subset of skills.
Given true $y \in \{0,1\}^6$ and prediction $\hat y$, we report
\emph{exact-match accuracy} (EM), which counts a prediction as correct
only when all six skill labels exactly match the ground-truth vector,
\emph{Hamming score} (fraction of correct predictions), and per-skill F1 score.

\subsection{Synthetic Data Pipeline}
\label{sec:data}

\begin{figure}[h]
\centering
\includegraphics[width=1\linewidth]{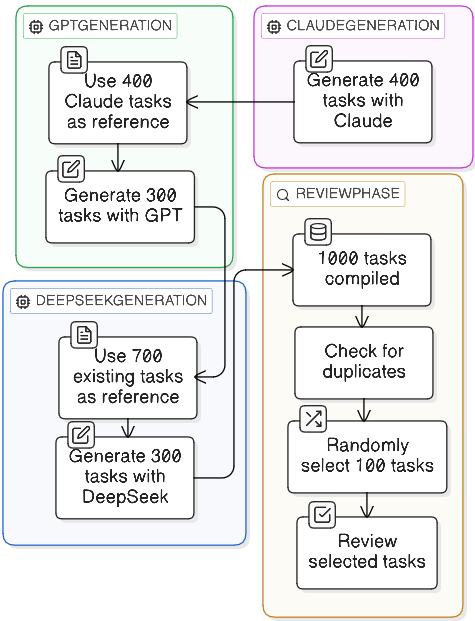}
\caption{Synthetic dataset creation pipeline.}
\label{fig:dataset-pipeline}
\end{figure}

\textbf{Task Diversity.}
We first generated 400 tasks with Claude Opus 4.5~\cite{claudeopus45}, then used that batch as context to generate 300 additional tasks with GPT-5~\cite{gpt5}. Using both earlier batches as references, we generated a final 300 tasks with DeepSeek-R1~\cite{deepseekr1}, for a total of 1{,}000 tasks (Figure~\ref{fig:dataset-pipeline}). We then randomly sampled 100 tasks for human review as a data quality sanity check. As a result, a set of tasks spans 22 application domains identified in advance:
agriculture, waterway operations, underwater/marine,
outdoor/wilderness, emergency/disaster, urban infrastructure, retail,
construction, warehouse/logistics, manufacturing, office/commercial,
energy/utilities, environmental/wildlife, medical, military/defense,
scientific research, residential, aerial operations, mining/geology,
aerospace/space, security/surveillance, and cross-domain generics. We randomly selected 100 tasks and manually verified their correctness.

\begin{figure}[t]
\centering
\includegraphics[width=0.9\linewidth]{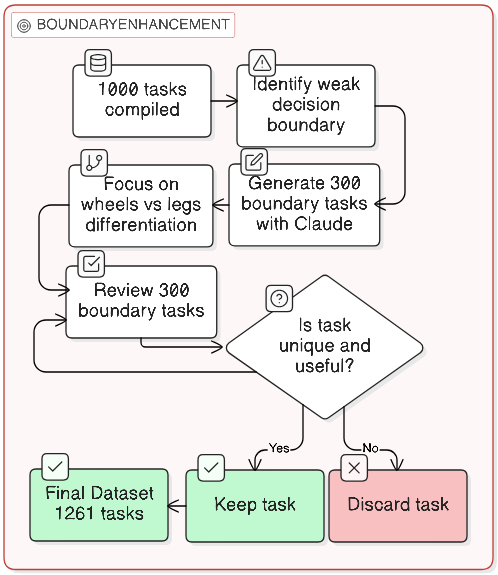}
\caption{Targeted boundary-task generation.}
\label{fig:boundary}
\end{figure}

\textbf{Targeted boundary-task generation.}
After training a first model (Section~\ref{sec:model}), per-skill F1 score exposes the weakest decision boundary. In our case, the
\texttt{legs} vs. \texttt{wheels} boundary dominated errors (22/34). We generated 300 additional tasks that are explicitly crafted to distinguish each side:
\texttt{legs}-only (rough terrain, no speed need), \texttt{wheels}-only (flat ground with speed or payload), \texttt{legs+wheels} (mixed-terrain missions that transition between surfaces), and similar disambiguation triples for the other weak boundaries (figure~\ref{fig:boundary}). After carefuly reviewing the data qualitiy of 300 additional tasks, 261 of 300 were added to our existing dataset. As a result, our final dataset consists of 1{,}261 tasks across 23 skill combinations with 1{,}061 training data and 200 test data.

\begin{table*}[t]
\centering
\small
\caption{Task-to-Skill Matching Performance on 200 /test Tasks}\label{tab:main}
\begin{tabular}{lcccc}
\toprule
\multicolumn{1}{c}{Model} & \multicolumn{1}{c}{Paramters ($\approx$)} & \multicolumn{1}{c}{Exact Match (\%)} & \multicolumn{1}{c}{Hamming Score (\%)} & \multicolumn{1}{c}{Macro F1 Score} \\
\midrule
Ensemble (distilled mpnet and MiniLM)      & 133M & \textbf{83.5} & \textbf{96.3} & \textbf{0.941} \\
all-mpnet-base-v2 (distilled)              & 110M & 81.5          & 96.0             & 0.931 \\
all-MiniLM-L6-v2 (distilled)              &  22M & 75.5          & 94.9             & 0.919 \\
\midrule
Kimi K2 (zero-shot)~\cite{kimik2}      & 1T / 32B$^\dagger$ & 72.0 & 93.8 & 0.887 \\
GPT-OSS (zero-shot)~\cite{gptoss} & 117B / 5.1B$^\dagger$ & 71.5 & 93.2 & 0.892 \\
Llama-4-Scout (zero-shot)~\cite{llama4} & 109B / 17B$^\dagger$ & 69.0 & 92.8 & 0.876 \\
\bottomrule
\end{tabular}\\[2pt]
\footnotesize{$^\dagger$ Total / active parameters (mixture-of-experts).}
\end{table*}

\subsection{Model and Training}
\label{sec:model}

\begin{table*}[!t]
\centering
\caption{Per-Skill F1 on 200 Test Tasks}
\label{tab:skill}
\small
\begin{tabular}{lrrrrrr}
\toprule
Skill & \shortstack{Ensemble\\(Distilled mpnet and MiniLM)} & \shortstack{all-mpnet-base-v2\\(distilled)} & \shortstack{all-MiniLML6-v2\\(distilled)} & \shortstack{Kimi-\\K2} & \shortstack{GPT-\\OSS} & \shortstack{Llama-4-\\Scout} \\
\midrule
Fly            & \textbf{0.981} & 0.971 & 0.952 & 0.936 & 0.898 & 0.929 \\
Legs           & 0.905 & \textbf{0.910} & 0.901 & 0.865 & 0.830 & 0.863 \\
Wheels         & \textbf{0.899} & 0.884 & 0.851 & 0.674 & 0.776 & 0.716 \\
Hands          & \textbf{0.943} & 0.941 & 0.898 & 0.933 & 0.870 & 0.835 \\
Under Water    & 0.939          & 0.902 & 0.958 & 0.958 & \textbf{0.979} & 0.958 \\
Surface Water  & 0.978          & 0.978 & 0.955 & 0.958 & \textbf{1.000} & 0.958 \\
\bottomrule
\end{tabular}\\[2pt]
\end{table*}

Each task is encoded by a sentence transformer~\cite{sentencetransformers}
(mean-pooled over token embeddings with attention-mask weighting) and
fed to a four-layer MLP head
with BatchNorm, GELU, and dropout between layers, producing six sigmoid logits. The encoder's last two transformer blocks are unfrozen during training; the remaining backbone stays frozen to prevent overfitting
on the small dataset. We train by distillating two encoders independently: \texttt{all-mpnet-base-v2}~\cite{mpnet}
($\approx$109M parameters, 12 layers, 768 dimensions and top two layers unfrozen) and \texttt{all-MiniLM-L6-v2}~\cite{minilm}
($\approx$23M parameters, 6 layers, 384 dimensions and top two layers unfrozen), then ensemble by averaging sigmoid probabilities
and thresholding at $0.5$.

\textbf{Hyperparameters.}
\texttt{BCEWithLogitsLoss} uses per-skill positive-class weights
($\mathrm{negative}/\mathrm{positive}$) to counter multi-label imbalance.
The full ensemble training is completes in on a Jetson
Orin AGX (GPU).

Per-skill thresholds tuned on a 15\% inner validation split of the training data produced slightly lower held-out EM than the fixed 0.5 threshold (80.5\% vs.\ 83.5\%), likely because the training distribution is not well matched to the boundary-heavy held-out set; we report the fixed-threshold ensemble as our primary system. Figure~\ref{fig:arch} illustrates the overall training and inference
pipeline of the ensemble model.

\section{Results}
\label{sec:results}
\subsection{Evaluation}
\textbf{Baselines:}
GPT-OSS-120B~\cite{gptoss} (reasoning Mixture of Experts (MoE), 117B total / 5.1B active parameters),
Llama-4-Scout-17B~\cite{llama4} (MoE, 16~experts, 17B active parameters / 109B total),
Kimi~K2~\cite{kimik2} (MoE, 1T total / 32B active parameters).
Our 133M-parameter ensemble model dominates on all metrics (exact match, Hamming score, and macro F1 score), beating the largest baseline (Kimi K2) by \textbf{+11.5\% EM} (Table~\ref{tab:main}). The individual trained encoders already beat the best
LLM baseline: all-mpnet-base-v2 scores 81.5\% EM and all-MiniLM-L6-v2 scores 75.5\% (Table~\ref{tab:main}).

Table~\ref{tab:skill} shows
per-skill F1. The largest gap is on \texttt{wheels}, driven by recall: all models predicted
\texttt{wheels} at only 52--63\% recall before the targeted boundary-task generation directly addressed this weakness. All systems handle the
water and flying skills easily because of unambiguous keywords (``underwater'', ``fly'', ``aerial'',
``river'') in task descriptions providing strong cues.

\textbf{Inference cost.} Each encoder processes tasks at roughly 10--20~ms/sample on a Jetson Orin AGX (CPU inference, no batching), vs.~6.4--7.0~s/sample for the Groq API~\cite{groq} baselines (median end-to-end latency, excluding rate-limit backoff), easily more than two orders of magnitude
faster per call, on top of the parameter-count gap.

\subsection{Prediction Analysis}
\label{sec:err}

Inspecting the LLM errors on the held-out set (200 tesk tasks) reveals two dominant failure modes.

\textbf{Wheels under-prediction.} Frontier LLMs predict \texttt{wheels} only when the description contains surface-mobility keywords (\emph{``drive''}, \emph{``flat surface''}, \emph{``patrol the floor''}), missing
tasks where speed or payload is \emph{implied} by context (``transport heavy machinery across a smooth factory floor'', ``rapid delivery between wards'', etc.). Our ensemble model significantly improved this obstacle by achiving 0.899 F1 score, a gap of atleast 0.123 compared to baseline models, which is a result of learning the boundary-task generation that explicitly contrasted ``needs speed'' vs.\ ``doesn't need speed'' on flat ground.

\textbf{Legs over-prediction on mixed-terrain tasks.} LLMs correctly output \texttt{legs} for tasks mentioning ``stairs'' or ``rough terrain'', but extend it to large flat outdoor spaces (``patrol the convention centre floor'', ``deliver meals across a hospital corridor'') where wheels alone suffice. This is an expected behavior given most environments that a robot with wheels can operate are also operatable by wheels.

\textbf{Underwater / surface water are saturated.} All systems,
including our ensemble, reach $\geq$0.90 F1 on these two skills
because the task descriptions contain strong lexical cues
(``underwater pipe'', ``surface of the lake'').

\section{Discussion}
Our results suggest that robot selection is better framed as a compact
skill-prediction problem than as a purely open-ended reasoning task for
large language models. More specifically, we introduce robot selection
from natural-language task descriptions as a distinct problem setting and
show that models can be trained to capture the relationship between task
requirements and robots' physical capabilities.

More broadly, this setting highlights a useful systems perspective for
future multi-robot assistants. In a heterogeneous fleet, the first
question is often not how to plan the task, but which robot should be
assigned to it. By separating robot selection from downstream planning,
our approach provides a simple interface between natural-language task
requests and fleet-level execution. As robot platforms and foundation
models continue to improve, such lightweight skill-prediction modules may
serve as a reliable front end for scalable robot routing in real-world
multi-robot deployments.

\textbf{Limitations.}
The main limitation of this study is data scale and label quality.
Although synthetic generation makes it possible to cover diverse task
domains efficiently, validating task descriptions and assigning correct
multi-label skill annotations still require careful human review. Our
manual verification of 100 sampled tasks provides an initial quality
check, but a larger and more systematic audit would strengthen confidence
in the dataset. In addition, the current benchmark focuses on a fixed
six-skill taxonomy, so the results do not yet capture finer-grained
capabilities such as payload limits, dexterity level, endurance, or
terrain-specific constraints.

\textbf{Future Work.}
A natural next step is to expand the dataset in both scale and complexity.
Beyond adding more task descriptions, we plan to incorporate operational
robot attributes such as autonomy (battery level and operable time), expected task
duration, and time-to-completion for different skill profiles. This would
move the problem from static skill matching toward more realistic robot
selection under resource and scheduling constraints. Another promising
direction is to extend the taxonomy beyond physical skills alone and
study how capability prediction can be combined with downstream planning
and fleet coordination in larger multi-robot systems.

\bibliographystyle{IEEEtran}
\bibliography{refs}

\end{document}